\documentclass{article}

\PassOptionsToPackage{numbers, sort&compress}{natbib}

\usepackage[preprint]{neurips_2021}
\usepackage[utf8]{inputenc}
\usepackage[T1]{fontenc}    

\usepackage{amsmath}
\usepackage{amsfonts}
\usepackage{nicefrac}       
\usepackage{mathtools}      

\usepackage[caption=false]{subfig} 
\usepackage{graphicx}
\usepackage{pgfplots}
\pgfplotsset{compat=1.15}
\usetikzlibrary{arrows}
\usetikzlibrary{backgrounds}
\usetikzlibrary{calc}
\usetikzlibrary{decorations.text}
\usetikzlibrary{decorations.markings}
\usetikzlibrary{fit}
\usetikzlibrary{intersections}
\usetikzlibrary{mindmap}
\usetikzlibrary{positioning}
\usetikzlibrary{shapes.geometric}
\usetikzlibrary{shapes}
\usepackage{tikzsymbols}
\usepackage{standalone}
\usepgfplotslibrary{groupplots}
\usepgfplotslibrary{fillbetween}
\usepackage{hyperref}

\usepackage{url}            
\usepackage{booktabs}       
\usepackage{microtype}      

\usepackage{xkeyval}
\usepackage{soul}
\usepackage[
    textsize=footnotesize,
    linecolor=blue,
    backgroundcolor=blue!25,
    bordercolor=blue,
    disable,
]{todonotes}

\usepackage[capitalise]{cleveref}

\usepackage{siunitx}

\newcommand{\thomsonlab}
{
    Biology and Biological Engineering\\
    California Institute of Technology\\
    Pasadena, CA 91125
}
\newcommand{\carnegiemellon}
{
    School of Computer Science\\
    Carnegie Mellon University\\
    Pittsburgh, PA 15213
}
\newcommand{\caltechmath}
{
    Applied and Computational Mathematics\\
    California Institute of Technology\\
    Pasadena, CA 91125
}

\usepackage{dsfont}
\usepackage{xspace}
\newcommand{\identity}{\ensuremath{\mathds{1}}\xspace}

\newcommand{\R}{\ensuremath{\mathbb R}\xspace}

\newcommand{\Np}{\ensuremath{N_p}\xspace}
\newcommand{\Ng}{\ensuremath{N_g}\xspace}
\newcommand{\Nt}{\ensuremath{N_t}\xspace}
\newcommand{\meas}{\ensuremath{\mathcal M}\xspace}
\newcommand{\diffvec}{\ensuremath{\vec d}}

\newcommand{\code}[1]{\texttt{#1}}

\newcommand{\timeseries}[5]{
    \includegraphics[width=.2\textwidth]{#1-#2.png}
    \includegraphics[width=.2\textwidth]{#1-#3.png}
    \includegraphics[width=.2\textwidth]{#1-#4.png}
    \includegraphics[width=.2\textwidth]{#1-#5.png}
}

\title{Reinforcement Learning reveals fundamental limits on the mixing of active particles}

\author{%
    Dominik Schildknecht\\\thomsonlab\\\texttt{dominik.schildknecht@gmail.com}
    \And
    Anastasia N.\ Popova\\\caltechmath
    \AND
    Jack Stellwagen\\\carnegiemellon
    \And
    Matt Thomson\\\thomsonlab\\\texttt{mthomson@caltech.edu}
}

\begin{document}

\maketitle

\begin{abstract}
    The control of far-from-equilibrium physical systems, including active materials, has emerged as an important area for the application of reinforcement learning (RL) strategies to derive control policies for physical systems. In active materials, non-linear dynamics and long-range interactions between particles prohibit closed-form descriptions of the system's dynamics and prevent explicit solutions to optimal control problems. Due to fundamental challenges in solving for explicit control strategies, RL has emerged as an approach to derive control strategies for far-from-equilibrium active matter systems. However, an important open question is how the mathematical structure and the physical properties of the active matter systems determine the tractability of RL for learning control policies. In this work, we show that RL can only find good strategies to the canonical active matter task of mixing for systems that combine attractive and repulsive particle interactions. Using mathematical results from dynamical systems theory, we relate the availability of both interaction types with the existence of hyperbolic dynamics and the ability of RL to find homogeneous mixing strategies. In particular, we show that for drag-dominated translational-invariant particle systems, hyperbolic dynamics and, therefore, mixing requires combining attractive and repulsive interactions. Broadly, our work demonstrates how fundamental physical and mathematical properties of dynamical systems can enable or constrain reinforcement learning-based control.
\end{abstract}

\section{Introduction} \label{sec:intro}
Mixing is studied both in abstract mathematics and in the context of applications. The mathematical treatment of mixing emerged in ergodic theory, which was used to understand how thermodynamics arises in deterministic systems~\cite{Boltzmann1896,Birkhoff1931,Neumann1932}. In addition, mixing applications are also crucial for various technologies at all length scales, ranging from large-scale chemical processes in industries such as petroleum or food~\cite{Paul2003} down to microscopic scales, including microfluidic applications in medicine~\cite{Valencia2012,Neuzil2012}. However, microscopic mixing is challenging because microscale systems are typically only weakly non-linear and drag-dominated, so efficient mixing strategies are not obvious. Current technical solutions for microscopic mixing and material transport rely on external pumps, prefabricated device geometries~\cite{Stroock2002,Gan2007,Kuo2011,Ortega-Casanova2016}, and additives~\cite{Steinberg2001}. In contrast, nature uses self-organization of interacting components to achieve microscopic transport and mixing more compact and energy-efficient~\cite{Guo2014}.

Recently, interest in replicating nature has emerged, and researchers have started to study so-called active matter, both to understand complex biological phenomena and to use it in applications. Here, active matter refers to systems of proteins or other particles that continuously consume energy to achieve non-equilibrium dynamics, which can self-organize to achieve specific tasks. In particular, recent work suggested using active matter to achieve macroscopic tasks, such as generating fluid flows~\cite{Marenduzzo2010,Qu2020a,Qu2021b}, flow rectification~\cite{Stenhammar2016}, and equilibration of glassy systems~\cite{Omar2019}. 

The unsolved scientific challenge now is how to assemble the existing (microscopic) systems to achieve macroscopic results. This question is challenging because systems are interaction-dominated, in contrast to other problems where agents or particles are self-propelled navigating through space, for example, in fluid dynamics~\cite{Reddy2016,Colabrese2017,Muinos-Landin2018,Reddy2018,Verma2018,Biferale2019,Schneider2019} or swarm robotics~\cite{Yu2010a,Rubenstein2014}. Because the system is interaction-dominated, strategies have to exert control over agent-agent interactions rather than single-agent properties. Due to this indirect control over individual agents, the control problem for active matter is more challenging compared to other physical environments. Therefore, finding successful strategies requires advanced control methods, including policy-based reinforcement learning.


The key question for this paper is how to find control strategies to steer large-scale particle systems with complex interactions to a desired mixing state. In particular, we search mixing strategies by applying RL to a drag- and interaction-dominated active matter model and observe that while RL fails to learn good strategies if only attractive or repulsive interactions are available, RL finds good strategies if attractive and repulsive interactions can be used. We analyze this puzzling behavior using dynamical systems theory, particularly theory to hyperbolic dynamics and Anosov diffeomorphisms, to prove that mixed interactions are indeed necessary to render the problem solvable. 

Broadly, in this paper, we show how the success of RL is determined by the mathematical structure of the dynamical system underlying the particles and their pairwise interactions. In particular, the main contributions of this paper are: (1) We develop an RL approach to microscopic interaction-dominated mixing. (2) The approach is applied to models with only attractive or repulsive interactions, where we observe failure to finding an efficient mixing strategy. In contrast, the same RL approach finds novel homogeneous mixing strategies for the model with combined attractive and repulsive interactions. (3) We use dynamical systems theory to prove the existence of a minimal instruction set to achieve mixing, demonstrating that RL is indeed expected only to succeed if both interactions are available to the agents.

\section{Related Work} \label{sec:literature}
The overarching goal for our work is to achieve homogeneous mixing using control over active components. While recently, machine learning was applied to various topics in active matter (see Ref.~\cite{Cichos2020} for a review), the literature on active matter control is limited. Microscopic active matter experiments and simulations demonstrated the ability to exert some control over fluid flow~\cite{Marenduzzo2010,Stenhammar2016,Norton2020,Qu2020a,Qu2021b},  global particle momentum~\cite{Falk2021}, and the accelerated equilibration of otherwise glassy systems~\cite{Omar2019}. In macroscopic systems, it was demonstrated that using control over single-agents, they can navigate through fluids at various Reynolds numbers~\cite{Reddy2016,Colabrese2017,Muinos-Landin2018,Reddy2018,Verma2018,Biferale2019,Schneider2019}, and self-organization can emerge in a multi-agent setting which can solve different macroscopic tasks~\cite{Yu2010a,Rubenstein2014,Yang2018,Zhang2019a,Durve2020}. In contrast to these papers, we analyze a more challenging system where we have merely control over pairwise interactions rather than single-particle properties, and we can connect the learning success of the RL algorithm with the mathematical structure of the dynamical system. We will thereby advance the current understanding of RL and its application to the control of interaction-dominated dynamical systems and additionally prove mathematically that the structure of the control problem can determine the success of RL.

In this paper, we search efficient control strategies to achieve homogeneous mixing in active matter systems. Mixing is typically studied as a part of ergodic theory, which lays the foundation of statistical physics, as it describes how thermodynamics arises from deterministic dynamical systems~\cite{Boltzmann1896,Birkhoff1931,Neumann1932}. In particular, (weak) mixing is a stronger form of ergodicity, i.e., ergodicity is a necessary but not sufficient condition for mixing (see textbooks~\cite{Walters1982,Kerr2016,Hawkins2021}). In contrast, the theory of dynamical systems shows that if a map is an Anosov diffeomorphism, it will induce (weak) mixing~\cite{Anosov1967,Brin2010}. 
In addition to the mathematical interest, mixing is also highly relevant in technical application, where it typically depends on the medium and the length scales involved. For example, in high Reynolds numbers fluids, mixing occurs due to advection~\cite{Batchelor1959,Dimotakis2000,Miles2018,Miles2018a}. 
At low Reynolds numbers, mixing is more challenging and requires the use of special geometries~\cite{Stroock2002,Gan2007,Kuo2011,Ortega-Casanova2016} or (active) additives~\cite{Steinberg2001,Saintillan2008,Saintillan2008a}, yet it has many applications, in particular in medical applications~\cite{Valencia2012,Neuzil2012}.
The findings of this paper will hopefully help design more efficient mixing devices for such applications.

\section{Environment, Rewards and Reinforcement Learning Framework} \label{sec:methods}
\subsection{Particle Mixing Environment} \label{sec:methods:env}
In this work, we will start by applying RL to a specific microscopic model of active matter. In particular, we want to select a sufficiently simple model to be amenable to RL while at the same time being suitable to describe various problems central to microscopic material transport and active matter. Microscopic models are often drag-dominated~\cite{Chaikin1995Ch7} and can be classified into continuum theories~\cite{Degennes1993,Joanny2007,Giomi2008,Saintillan2008,Saintillan2008a,Giomi2015,Foster2015a,Qu2021b} and particle-based models~\cite{Berendsen1995,Plimpton1995,Nedelec2007,Schildknecht2020}. While continuum theories are often preferred due to their analytic properties, we chose a particle-based model because they tend to be quicker to simulate small- and medium-sized systems, and therefore more amenable to RL requiring many simulations. In addition, since we are interested in general properties, a phenomenological model is sufficient, and we settled on the simulation platform proposed by Ref.~\cite{Schildknecht2020}, used to simulate active matter systems~\cite{Ross2019,Qu2020a,Qu2021b}. It should be noted that our results will be generalized to a larger array of models in \cref{sec:analytics}.

The model introduced by Ref.~\cite{Schildknecht2020} required only attractive interactions to describe the existing experimental platform. However, this restriction will turn out to limit mixing. Hence, an extended model is introduced as follows: The system consists of \Np particles in two dimensions indexed by $i$, which can be in three states: attractive-activated ($p_i^a=1,~p_i^r=0$), repulsive-activated ($p_i^a=0,~p_i^r=1$), or inactivated ($p_i^a=0,~p_i^r=0$). Particles can be activated by being in the corresponding activation area (which will form our control input) and deactivate with a rate $\lambda$ outside of activation areas. Activated particles interact with other similar-activated particles by a spring potential.\footnote{I.e., attractive-activated particles only interact with other attractive-interacted particles, and repulsive-activated particles only interact with other repulsive-activated particles.} The system's dynamics is described by the equations of motion
\begin{subequations}
    \begin{align}
        \vec x_i(t+\Delta t) &=  \vec x(t)+\frac{\Delta t^2}{m}\vec F_i,\label{eq:eom:x-and-F}&
        \vec F_i &= \vec F_i^{a}+\vec F_i^{r},\\
        \vec F_i^{a} &= -\nabla_{i} \Big[ \frac{k}{2}\sum_{\mathclap{\substack{j\neq i\\r_c<\,|\vec x_i-\vec x_j|\,<R_c}}}p^a_i p^a_j|\vec x_i-\vec x_j|^2 \Big], \label{eq:eom:att-and-rep}&
        \vec F_i^{r} &= -\nabla_{i} \Big[ \frac{k}{2}\sum_{\mathclap{\substack{j\neq i\\r_c<\,|\vec x_i-\vec x_j|\,<R_c}}}p^r_i p^r_j(|\vec x_i-\vec x_j|-R_c)^2 \Big],
    \end{align}
    \label{eq:eom}%
\end{subequations}
where $\vec x_i(t)$ is the position of particle $i$ at time $t$, and $\vec F_i$ is the total force acting on it. The total force $\vec F_i$ is split into attractive and repulsive contributions, and at every times step, at least one of them vanishes because at least one of $p_{i}^a$ and $p_{i}^r$ is $0$. It should be noted that the attractive and the repulsive force incorporate the same spring constant $k$, lower truncation $r_c$, and upper truncation $R_c$, and differ only by their rest length being $0$ or $R_c$ for the attractive and repulsive force, respectively. 

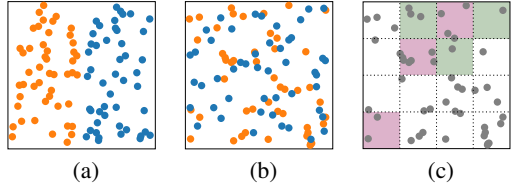
\begin{figure}
    \centering
    \subfloat[][]{
        \resizebox{.15\linewidth}{!}{ \begin{tikzpicture}
    \definecolor{C0}{HTML}{1F77B4}
    \definecolor{C1}{HTML}{ff7f0e}

    \draw[black] (0,0) rectangle (4,4);

    \pgfmathsetseed{42}
    \foreach \x in {1,...,48}
    {
      \pgfmathrandominteger{\a}{10}{190}
      \pgfmathrandominteger{\b}{10}{390}

      \pgfmathparse{0.01*\a}\let\a\pgfmathresult
      \pgfmathparse{0.01*\b}\let\b\pgfmathresult

      \fill[C1] (\a,\b) circle (0.10);
    };
    \foreach \x in {1,...,48}
    {
      \pgfmathrandominteger{\a}{210}{390}
      \pgfmathrandominteger{\b}{10}{390}

      \pgfmathparse{0.01*\a}\let\a\pgfmathresult
      \pgfmathparse{0.01*\b}\let\b\pgfmathresult

      \fill[C0] (\a,\b) circle (0.10);
    };
\end{tikzpicture} }
        \label{fig:sketch:initial}
    }
    \subfloat[][]{
        \resizebox{.15\linewidth}{!}{ \begin{tikzpicture}
    \definecolor{C0}{HTML}{1F77B4}
    \definecolor{C1}{HTML}{ff7f0e}

    \draw[black] (0,0) rectangle (4,4);

    \pgfmathsetseed{23}
    \foreach \x in {1,...,48}
    {
      \pgfmathrandominteger{\a}{10}{390}
      \pgfmathrandominteger{\b}{10}{390}

      \pgfmathparse{0.01*\a}\let\a\pgfmathresult
      \pgfmathparse{0.01*\b}\let\b\pgfmathresult

      \fill[C1] (\a,\b) circle (0.10);
    };
    \foreach \x in {1,...,48}
    {
      \pgfmathrandominteger{\a}{10}{390}
      \pgfmathrandominteger{\b}{10}{390}

      \pgfmathparse{0.01*\a}\let\a\pgfmathresult
      \pgfmathparse{0.01*\b}\let\b\pgfmathresult

      \fill[C0] (\a,\b) circle (0.10);
    };
\end{tikzpicture} }
        \label{fig:sketch:target}
    }
    \subfloat[][]{
        \resizebox{.15\linewidth}{!}{ \begin{tikzpicture}
    \definecolor{C0}{HTML}{1F77B4}
    \definecolor{C1}{HTML}{ff7f0e}
    \definecolor{CP}{HTML}{8e0152} 
    \definecolor{CG}{HTML}{276419} 

    \fill[CP!30] (2,3) rectangle ++ (1,1);
    \fill[CP!30] (1,2) rectangle ++ (1,1);
    \fill[CP!30] (0,0) rectangle ++ (1,1);
    \fill[CG!30] (2,2) rectangle ++ (1,1);
    \fill[CG!30] (3,3) rectangle ++ (1,1);
    \fill[CG!30] (1,3) rectangle ++ (1,1);

    \draw[black] (0,0) rectangle (4,4);

    \pgfmathsetseed{23}
    \foreach \x in {1,...,48}
    {
      \pgfmathrandominteger{\a}{10}{390}
      \pgfmathrandominteger{\b}{10}{390}

      \pgfmathparse{0.01*\a}\let\a\pgfmathresult
      \pgfmathparse{0.01*\b}\let\b\pgfmathresult

      \fill[black!50] (\a,\b) circle (0.10);
    };

    \foreach \x in {1,2,3}{
        \draw[thick, dotted] (0,\x) -- ++ (4,0);
        \draw[thick, dotted] (\x,0) -- ++ (0,4);
    }

\end{tikzpicture} }
        \label{fig:sketch:binning}
    }
    \caption{Sketch of the setup: In \cref{fig:sketch:initial}, the initial condition is depicted with orange ``left''-tagged particles and blue ``right''-tagged particles. In \cref{fig:sketch:target}, a good solution that we would like to achieve is depicted using the same color-coding. In \cref{fig:sketch:binning}, we demonstrate the binning and color-code a possible activation pattern with attractive-activation areas depicted in green, repulsive-activation areas in pink and non-activating areas in white.}
\end{figure}

Using this model of drag-dominated particle dynamics with controllable pairwise spring interactions, we built an RL environment using OpenAI gym~\cite{Brockman2016}. In particular, $\Np=96$ particles are initialized randomly in a box with periodic boundary conditions, asserting an equal number of particles on the two halves of the system (for future reference, tagged ``left'' and ``right''). A sketch of the initial condition can be found in \cref{fig:sketch:initial}, and an example of a target state is depicted in \cref{fig:sketch:target}. The system is then integrated according to \cref{eq:eom} for $\Nt=100$ time steps. 
The detailed simulation parameters can be found in the supplemental material A.1. The observation and control spaces are introduced by using an $\Ng\times \Ng=4\times 4$ square grid as depicted in \cref{fig:sketch:binning}. In particular, observations are given to the algorithm by providing a separate count for each tag (corresponding to the orange and blue colors in \cref{fig:sketch:initial,fig:sketch:target}) and each bin. 
The system is controlled by associating each square of the binning to either activation area or none at all so that the action space at every time step is $3^{4\times 4}\approx 43\cdot 10^6$ dimensional if both interactions are included.

\subsection{Combined Mixing and Homogeneity Rewards} \label{sec:methods:reward}
To apply RL, we need to measure how well mixed a given state is, i.e., we need to define a functional giving a configuration as depicted in \cref{fig:sketch:initial} a low score and one as in \cref{fig:sketch:target} a high score. Various approaches could be taken to define such a function: One could use variations of the continuous mix-norm~\cite{Doering2006,Thiffeault2012,Miles2018,Miles2018a} or use an adversarial neural network approach to distinguish a well-mixed state from a segregated state. However, as we will demonstrate in \cref{sec:analytics}, the main results of our paper will be unaffected by the choice of the reward function, and hence, the compute-intensive reward functions can be replaced by a simpler one. In particular, we reuse the input tensor \meas, i.e., the count of particles with a specific tag per bin, which has the indices (tag, $x$-bin, $y$-bin), where tag $\in\{l,r\}$, and $x$-bin and $y$-bin $\in\{1, 2,\ldots,\Ng\}$. Then, we define the mixing reward as
\begin{align}
    \mathcal R_m&= -\frac{1}{\Nt\,\mathcal N_m} \sum_{x=1}^{\Ng}\sum_{y=1}^{\Ng} \left( \meas_{l, xy}-\meas_{r, xy} \right)^2\in \left[ -\frac{\Ng^2}{2},0 \right],
    \label{eq:reward:mixing}
\end{align}
where the minus sign ensures that the reward is larger for well-mixed states and $\mathcal N_m=({\Np}/{\Ng})^2$ is chosen so that the reward for the initial state is $-1$ on average (if no action is taken).

Since only using \cref{eq:reward:mixing} will lead to degenerate solutions that collapse all particles to dense clusters, we add an additional homogeneity reward, constructed analogously to \cref{eq:reward:mixing}:
\begin{align}
    \mathcal R_h= -\frac{1}{\Nt\,\mathcal N_h} \sum_{x=1}^{\Ng}\sum_{y=1}^{\Ng}
                \left[ 
                    \frac{\Np}{\Ng^2} 
                    -(\meas_{l,xy}+\meas_{r,xy})
                \right]^2\in [-1,0],
    \label{eq:reward:homogeneity}
\end{align}
but rather than comparing the particle counts per tag in each bin, the homogeneity reward compares the number of particles in each bin $(\meas_{l,ij}+\meas_{r,ij})$ to the average number of particles $\frac{\Np}{\Ng^2}$ per bin. Here, $\mathcal N_h=\Np^2\left( 1-\frac{1}{\Ng^2} \right)$ so that the most inhomogeneous solution, namely all particles in a single bin, has a reward of $-1$. Finally, the two reward contributions are combined to 
    $\mathcal R = \alpha \mathcal R_m+ (1-\alpha) \mathcal R_h$
using the parameter $\alpha\in [0,1]$ to tune between weighting more homogeneous or more mixed solutions continuously.

\subsection{Reinforcement Learning Framework} \label{sec:methods:rl}
Because the action space is large ($3^{4\times 4}\approx 43\cdot 10^6$ for attractive and repulsive interactions), conventional control theory techniques are intractable due to Bellman's curse of dimensionality~\cite{Bellman1957}. Similarly, the action space is also too large for value-based RL methods, so that we use a policy-based algorithm. In particular, we use the \code{rllib} implementation~\cite{Liang2017} of Proximal Policy Optimization (PPO)~\cite{Schulman2017}, with a neural network consisting of three convolutional layers (with 32, 64, 256 channels, respectively, all with a kernel size of 3), and a dense layer at the end. The hyperparameters for PPO were tuned automatically during training using \code{ray tune}'s Population Based Training (PBT)~\cite{Liaw2018} with 16 agents. Using PBT, unstable solutions could be overcome, and we observed converged strategies in all training situations. Additional hyperparameters required for PBT can be found in the supplemental material A.2. 

\section{Reinforcement Learning Training Results} \label{sec:results}
\subsection{Attractive Interactions lead to a Collapse} \label{sec:results:attractive}
\begin{figure}
    \centering
    \includegraphics[width=\linewidth]{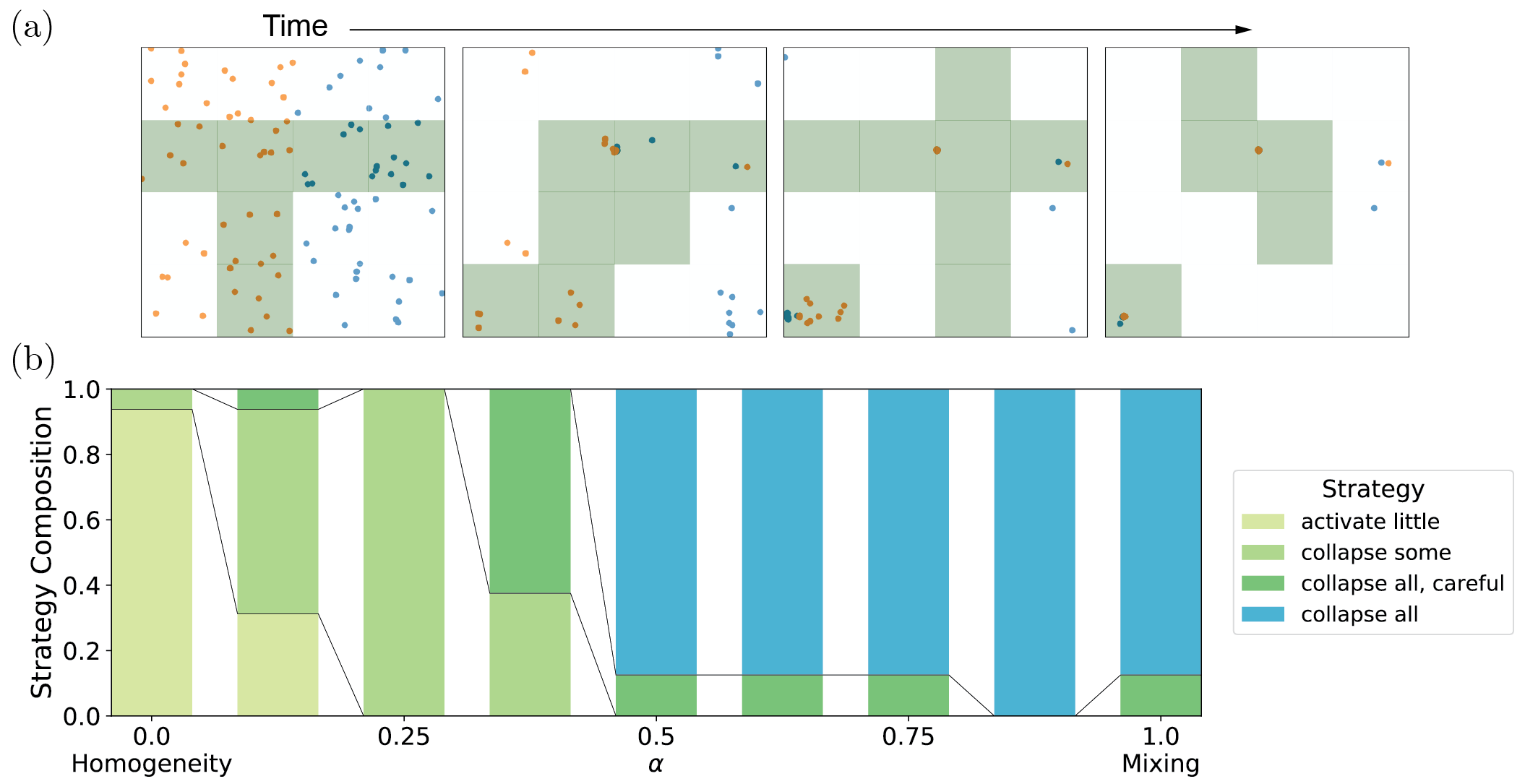}
    \caption{Training with only attractive interactions achieves only collapsed clusters: In \cref{fig:attractive}a, an exemplary time series for only mixing-focused training ($\alpha=1$) is shown, in which a collapse of the particle density can be observed. In \cref{fig:attractive}b, the composition of emergent strategies at various values of $\alpha$ is shown, where all the intermediate emergent strategies are taking over continuously. However, no emergent strategy displays a homogeneous mixing.}
    \label{fig:attractive}
\end{figure}

Experimentally, a controllable active matter system with attractive interactions only is already available~\cite{Ross2019,Qu2020a,Qu2021b}, so this case is analyzed first. An exemplary time series for a converged strategy for the mixing focused $\alpha=1$ case is shown in \cref{fig:attractive}a, where the emergent strategy collapses all particles into a dense cluster. While this strategy is not a good mixer in the common sense, the solution optimizes \cref{eq:reward:mixing}. In particular, if all particles collapse to a cluster in a single bin, then this bin has a balanced number of ``left'' and ``right'', so that for this cell $\meas_{l, ij}=\meas_{r, ij}=\Np/2$, and the difference is $0$. All other cells are empty so that for these cells $\meas_{l, ij}=\meas_{r, ij}=0$ again leading to a vanishing difference, and hence $\mathcal R=0\cdot\mathcal R_h + 1\cdot\mathcal R_m=0$ is optimal.

To overcome the trivial ``collapse all''-solution, $\alpha$ can be reduced so that homogeneity is weighted in. For the extreme case of $\alpha=0$, where the algorithm optimizes for homogeneity only, almost no particles were activated as the initial condition already has a high homogeneity. Indeed, also for intermediate values of $\alpha$, no efficient mixing strategies emerge. In order to see this, we hand-labeled the last validation video for each of the 16 PBT agents and plotted the emergent strategy composition as a function of $\alpha$ in \cref{fig:attractive}b. While a continuous change over multiple strategies emerges, none achieves a homogeneous mixed state.

Therefore, using only attractive interactions, no strategy emerged that provides a solution similar to our target, as depicted in \cref{fig:sketch:target}, even when varying $\alpha$. Indeed, we will show in \cref{sec:analytics} that no reward-shaping could overcome this problem, as attractive interactions are insufficient to arrive at a homogeneous and mixed state. 

\subsection{Only using repulsive Interactions achieves mixing by exploiting Periodic Boundary Conditions} \label{sec:results:repulsive}
\begin{figure}
    \centering
    \includegraphics[width=\linewidth]{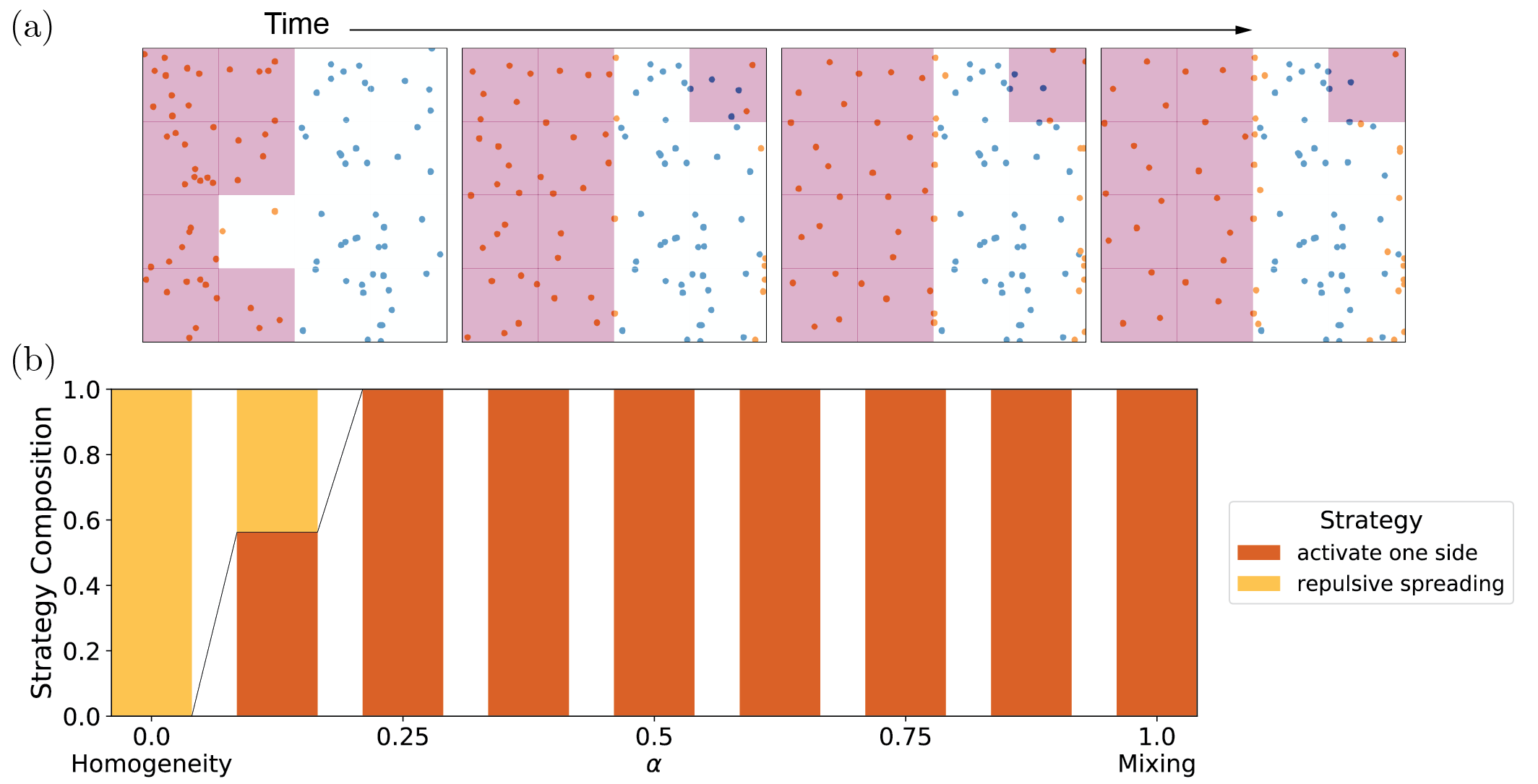}
    \caption{Training with repulsive interactions has a dominant strategy using the boundary conditions to achieve mixing: In \cref{fig:repulsive}a, an exemplary time series for only mixing-focused training ($\alpha=1$) is shown, in which activation of one side of the system can be observed, using the periodic boundary conditions to facilitate mixing. In \cref{fig:repulsive}b, the different emergent strategies are shown, demonstrating that the strategy is dominant over most of the $\alpha$-parameter space.}
    \label{fig:repulsive}
\end{figure}
While not yet implemented in the experimental platform, a repulsive-interactions-only system should be feasible to engineer, and there exists a more homogeneous mixing strategy compared to the attractive-only case: A typical time series for the mixing focused $\alpha=1$ case is shown in \cref{fig:repulsive}a, where we observe that the emergent strategy is the ``activate one side''-strategy. Due to the periodic boundary conditions, every particle expelled from one side enters the other side without contracting to a cluster. Two features of this mixing solution should be highlighted: First, this strategy again maximizes $\mathcal R_m$ by reducing the homogeneity of the system. Second, mixing is only facilitated by exploiting the periodic boundary conditions. 

For $\alpha=0$, i.e., maximizing the homogeneity of the system, the network uses repulsive interactions in more densely populated bins to achieve homogenization beyond the initial condition (``repulsive spreading''-strategy). However, as becomes clear from \cref{fig:repulsive}b, this spreading strategy only emerges very close to $\alpha=0$ and is quickly overtaken by the ``activate one side'' strategy. Indeed, the homogeneity reward strongly punishes dense clusters as emergent in the ``collapse all''-strategy but only weakly punishes a situation where only half of the cells become empty. Hence, the ``activate one side''-strategy is dominant over large parts of the $\alpha$-parameter-space. While the ``activate one side''-strategy achieves some mixing with the repulsive-only interactions, the strategy is not tunable, and it heavily relies on the periodic boundary conditions. Indeed, we will demonstrate in \cref{sec:analytics} that phase-space-limiting boundary conditions (such as periodic boundary conditions) are required to facilitate mixing with repulsive interactions only.

\subsection{Including attractive and repulsive Interactions enables diverse Strategies} \label{sec:results:both}
\begin{figure}
    \centering
    \includegraphics[width=\linewidth]{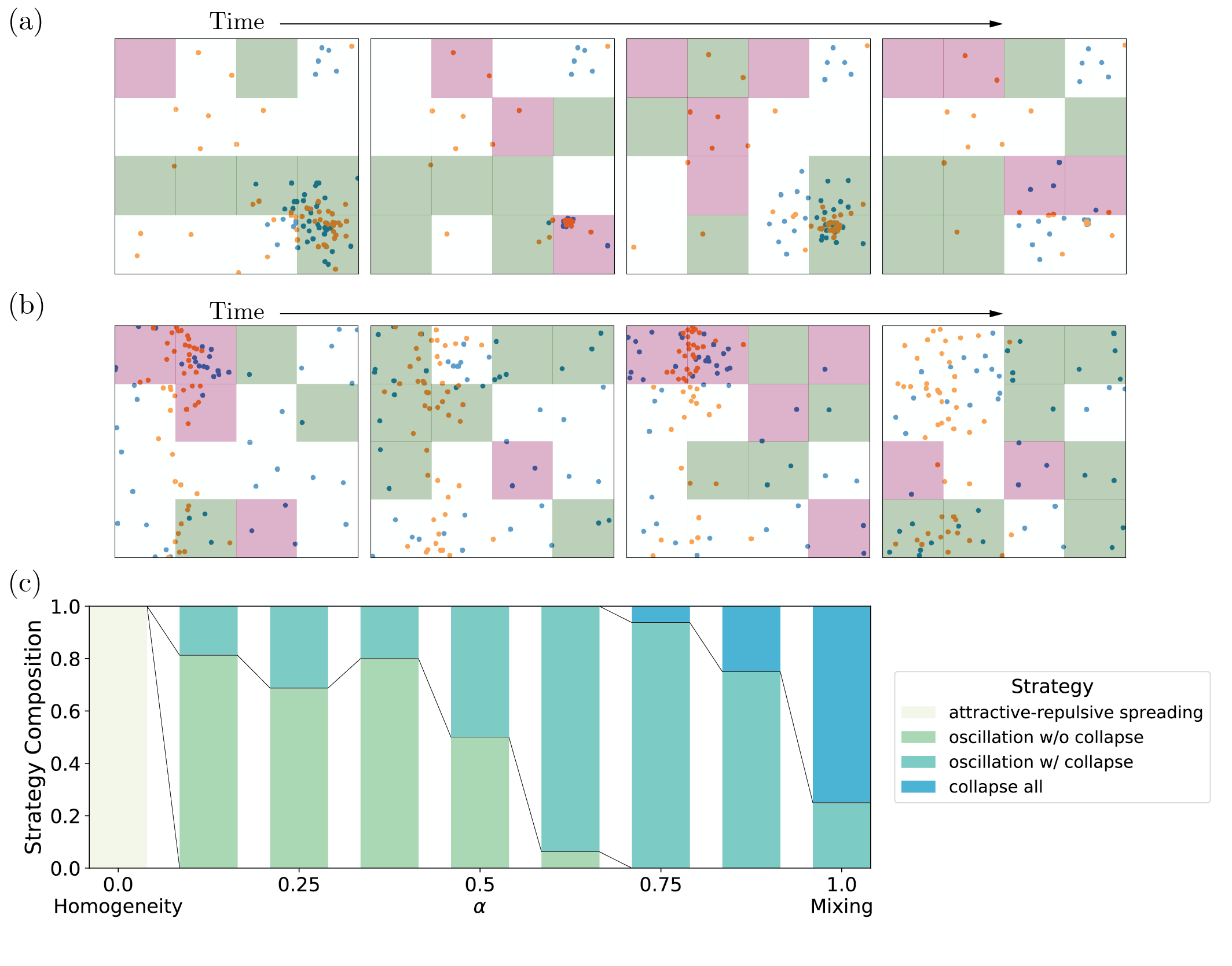}
    \caption{Training on a system with both interactions achieves homogeneous mixing by alternating attractive and repulsive interactions: In \cref{fig:both}a and b, two exemplary time series for oscillation strategies are shown, with and without collapse, respectively. In \cref{fig:both}c, the fraction of runs with a particular strategy at various values of $\alpha$ is shown, with various emergent strategies continuously transforming into each other.}
    \label{fig:both}
\end{figure}

In contrast to the previous cases, using both interaction types, we will achieve a tunable mixing strategy that can achieve both homogeneity and mixing without exploiting the boundary conditions. Starting with the $\alpha=1$ case, we observe the ``collapse all''-strategy emerging once more. However, already for $\alpha=1$, the network sometimes uses repulsive interactions to spread out dense particles, only to attract them again. Because this cycle typically occurs multiple times, we call these strategies oscillatory, and they emerge in different flavors. For large $\alpha$, the emergent strategy still collapses particles to a dense cluster (see \cref{fig:both}a) in a strategy we term ``oscillation w/ collapse''. For smaller $\alpha$, the strategies still try to use oscillations but avoid very dense clusters in a strategy we call ``oscillations w/o collapse'', depicted in \cref{fig:both}b. Finally, for $\alpha=0$, the RL algorithm finds an ``attractive-repulsive spreading''-strategy which homogenizes the system using both interaction types to spread dense cells and collapse in less populated cells. 

Overall, an interesting strategy evolution as a function of $\alpha$ emerges using both interactions, as shown in \cref{fig:both}c. In particular, a smooth transition from one strategy to the next can be observed, and various intermediate strategies produce well-mixed states without relying on boundary conditions. This result is insofar surprising as attractive or repulsive interactions alone are insufficient to produce efficient mixing, but the combination of both is sufficient. We can now analyze the success-failure pattern of RL to find efficient strategies using dynamical systems theory to analyze the problem more closely. Doing so will reveal that the necessity of both interactions is not a specific feature of the model in \cref{eq:eom} but a much more general statement about different models.

\section{Dynamical Systems Theory reveals the necessity of attractive and repulsive Interactions} \label{sec:analytics}
In the numerical results presented in \cref{sec:results}, we observed that only the combination of attractive and repulsive interactions led to homogeneously mixed states and that tweaking the reward function did not find a suitable solution for restricted interaction sets. While these results are limited in their applicability to the model described by \cref{eq:eom}, here, we will demonstrate how these results can be generalized using dynamical system theory and ergodic theory. In particular, we will show that homogeneous mixing in drag-dominated translational-invariant particle systems can only occur in systems with attractive and repulsive interactions.

We start by reconsidering the model in \cref{eq:eom}. Without truncating the sums, the equations of motion are
\begin{align}
    \vec x_i (t+\Delta t) 
    &= \vec x_i(t)-\frac{kp_i\Delta t^2}{m}\sum_{\substack{i\neq j\\p_i=1=p_j}}\frac{|\diffvec_{ij}(t)|-r_0}{|\diffvec_{ij}(t)|} \diffvec_{ij}(t),
    \label{eq:combined:eom}
\end{align}
where $r_0=0$ in the case of attractive interactions, and $r_0=R_c$ for repulsive interactions. Here, the difference vector is abbreviated as $\diffvec_{ij}(t)=\vec x_i(t)-\vec x_j(t)$. Hence, the updates without periodic boundary conditions can be written as a matrix-vector multiplication: $\vec X(t+\Delta t)=M\vec X(t)$, where $\vec X\in\R^{2\Np}$ is the collection of all particle positions. It should be noted that $M$ depends on which particles are activated (hence making it time-dependent), and on $\vec X$ (making the update non-linear). However, both dependencies are small since $M{[\vec X, t]}=\identity +\Delta t^2 \tilde M{[\vec X, t]}$, where $\Delta t\ll 1$. 

The long-term evolution of the system can be analyzed by considering the eigenvalues $\lambda_i$ of $M$ at every time step. Namely, ergodicity (and therefore mixing) requires the update to be measure-preserving in phase space~\cite{Hawkins2021}, meaning that the determinant $\det M = \prod_i \lambda_i$ has to be 1. If the update map $M$ is sufficiently differentiable, has eigenvalues smaller and larger than $1$ but maintains a determinant of $1$, then the map is so-called Anosov and is guaranteed to mix~\cite{Anosov1967,Brin2010}. 

Therefore, if we understand the spectrum of $M$ (i.e., the set of all eigenvalues), strong statements about mixing can be achieved. To compute the spectrum, it is sufficient to consider $M$ being represented as
\begin{align}
    M&= \begin{pmatrix}
          {1-\sum_{j}}c_{1j}           & \phantom{1-\sum_{j}}c_{12} & \phantom{1-\sum_{j}}\dots         & \phantom{1-\sum_{j}}c_{1N_a}\\
          \phantom{1-\sum_{j}}c_{21}   & \phantom{1-\sum_{j}}\ddots & \phantom{1-\sum_{j}}\ddots        & \phantom{1-\sum_{j}}\vdots\\
          \phantom{1-\sum_{j}}\vdots   & \phantom{1-\sum_{j}}\ddots & \phantom{1-\sum_{j}}\ddots        & \phantom{1-\sum_{j}}c_{N_a-1,N_a}\\
          \phantom{1-\sum_{j}}c_{N_a1} & \phantom{1-\sum_{j}}\dots  &\phantom{1-\sum_{j}} c_{N_a,N_a-1} & 1-\sum_{j}c_{N_aj},
      \end{pmatrix},
    \label{eq:structure-M}
\end{align}
where $|c_{ij}|\ll 1$ because $c_{ij}\propto \Delta t^2$. Here, it should be noted that such a form of $M$, namely one with row sum $1$ and small off-diagonal elements, will generally arise for particle-based drag-dominated translational-invariant systems~\cite{Chaikin1995Ch7}, hence all the following results are valid for a large class of models. Additionally, attractive interactions state that $c_{ij}\geq 0$, and for repulsive interactions $c_{ij}\leq 0$. These properties are sufficient to make a strong statement about the eigenvalues of $M$. 

Because $M$ is symmetric, all eigenvalues are real.\footnote{For non-symmetric matrices, the argument would be analogous, limiting the discussion to the absolute value of the eigenvalue.} Because of Gershgorin's circle theorem~\cite{Gershgorin1931}, the spectrum of $M$ 
\begin{align}
    \lambda(M)\subset\bigcup_{i} \left[ 1-\sum_{j\neq i} c_{ij}-\sum_{j\neq i}|c_{ij}|, 1-\sum_{j\neq i} c_{ij}+\sum_{j\neq i}|c_{ij}| \right].
    \label{eq:gershgorin}
\end{align}
Using \cref{eq:gershgorin} for attractive interactions where $c_{ij}\geq 0$: $\lambda(M)\subset [1-2\max_i\sum_{j\neq i}c_{ij}, 1]$ so that the largest possible eigenvalue is $1$, while all other eigenvalues are less than $1$. Therefore, $\det M\leq 1$ and iterating $M$ on any initial vector will either collapse the phase space direction (associated eigenvalue $\lambda<1$) or leave it invariant (associated eigenvalue $\lambda=1$). Hence, with only attractive interactions, the system will only be able to collapse, never leading to homogeneous mixing.

Analogously, for repulsive interactions, where $c_{ij}\leq 0$, $\lambda(M)\subset [1,1+2\max_i\sum_{j\neq i}|c_{ij}|]$ so that all eigenvalues are larger or equal to $1$, and $\det M\geq 1$. Therefore, without phase-space-limiting boundary conditions, the phase space is expanding, never leading to homogeneous mixing. However, with periodic boundary conditions, phase space is limited and can be folded into itself. Thus, the successful mixing with only repulsive interactions presented in \cref{sec:results:repulsive} was only possible due to the periodic boundary conditions. 

Finally, for attractive and repulsive interactions, the off-diagonal elements are around $0$ but can have either sign\footnote{Per row, each off-diagonal will have the same sign, as only same activation states interact with each other. However, due to the possibility of changing the interaction state, the same row can have an opposite sign at different times.} so that the eigenvalues of $M$ lie around $1$ (more precisely $\lambda(M)\subset [1-2\max_i\sum_{j\neq i}|c_{ij}|,1+2\max_i\sum_{j\neq i}|c_{ij}|]$). While it is not guaranteed that the determinant will be $1$, the updates now can induce hyperbolic dynamics, i.e., phase space stretching in some directions and compressing in others, bringing the dynamics closer to an Anosov diffeomorphism, hence being able to induce mixing in agreement with our successful mixing simulations in \cref{sec:results:both}. Our results furthermore indicate that in order to optimize mixing, the determinant should be close to $1$, which could be achieved in future work by introducing an additional reward or fine-tuning $\alpha$.

\begin{figure}
    \centering
    \subfloat[][Attractive interactions]{
        \includegraphics[width=.3\linewidth]{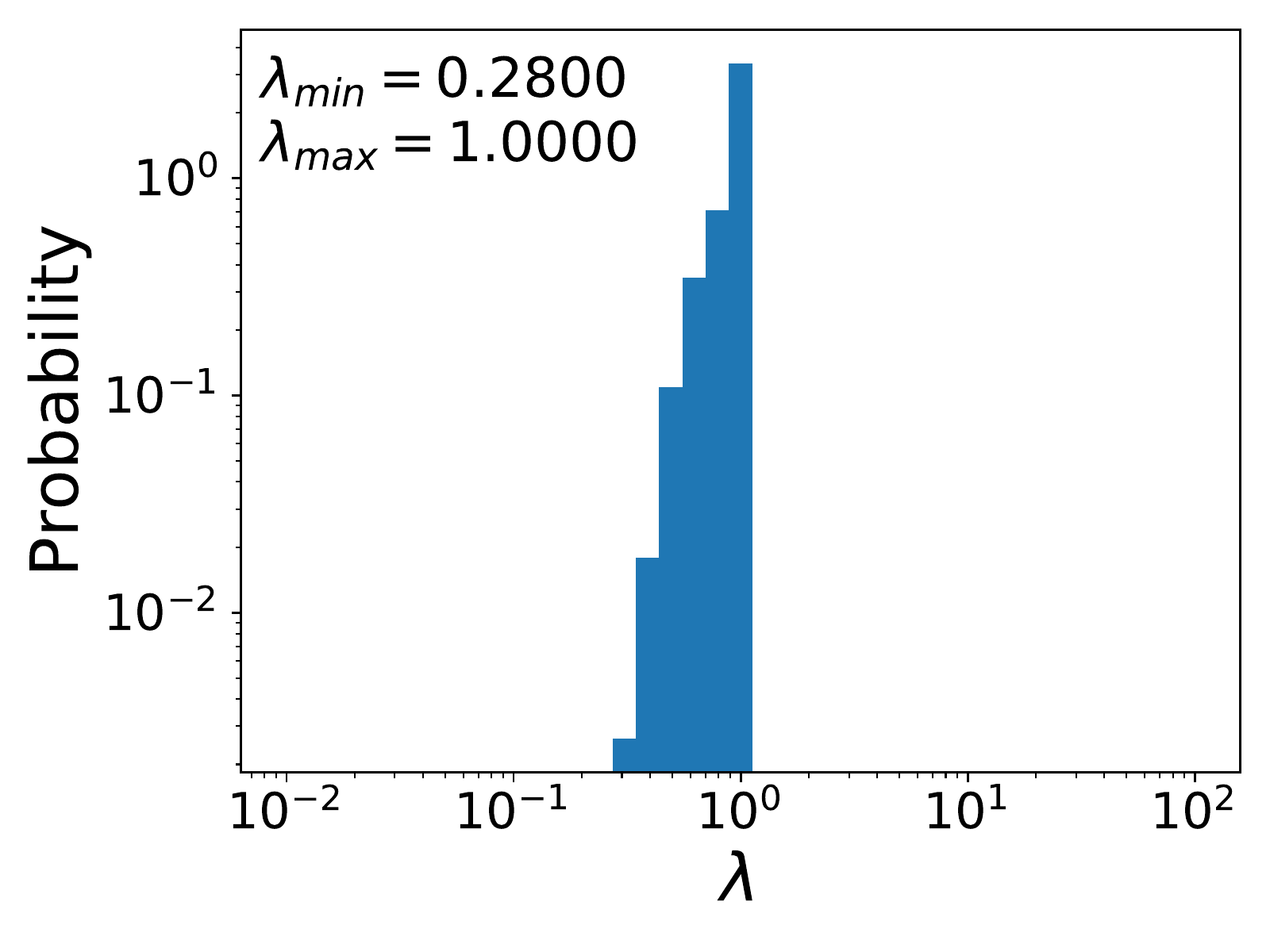}
        \label{fig:ev:att}
    }
    \subfloat[][Repulsive interactions]{
        \includegraphics[width=.3\linewidth]{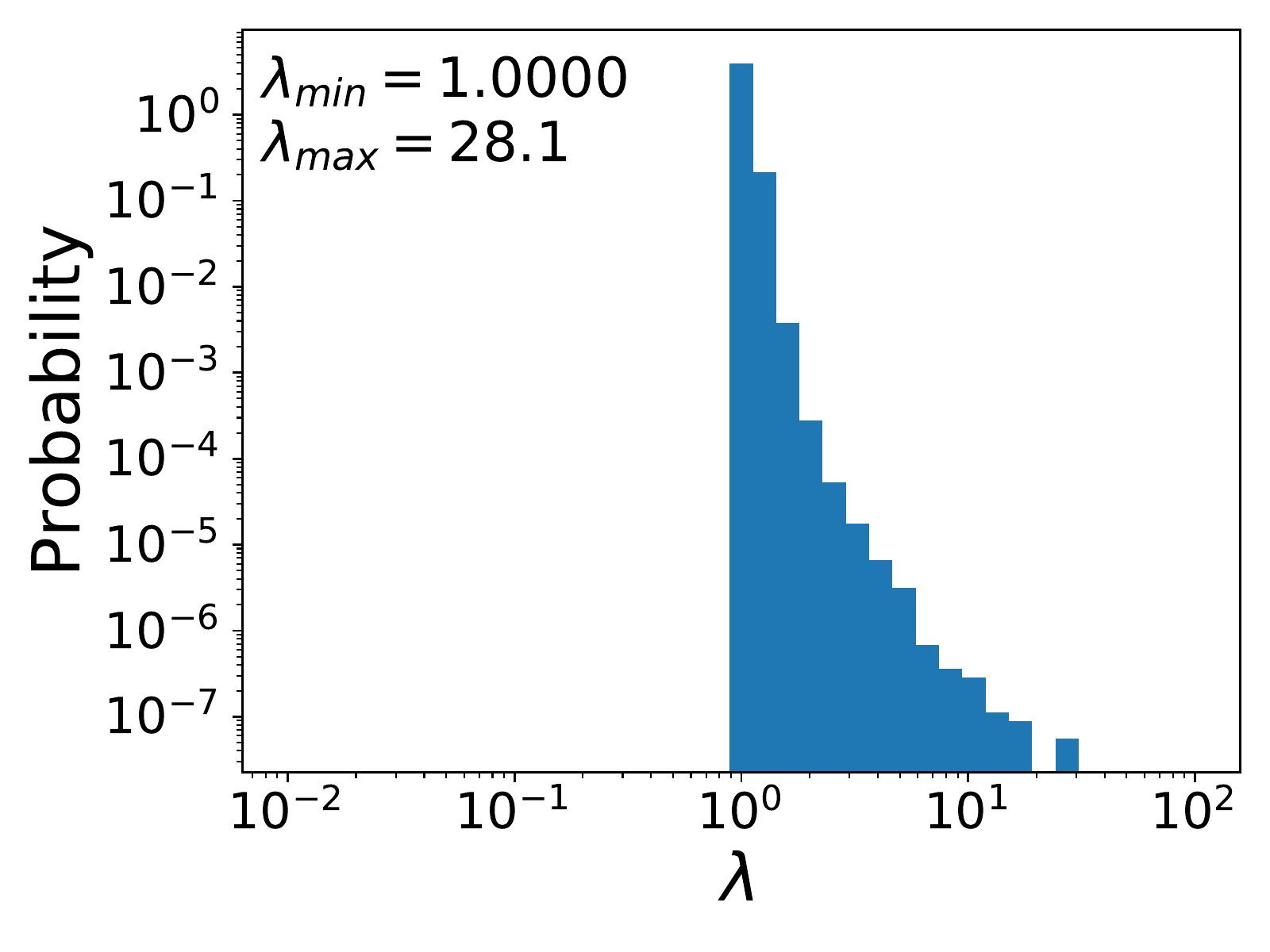}
        \label{fig:ev:rep}
    }
    \subfloat[][Both interactions]{
        \includegraphics[width=.3\linewidth]{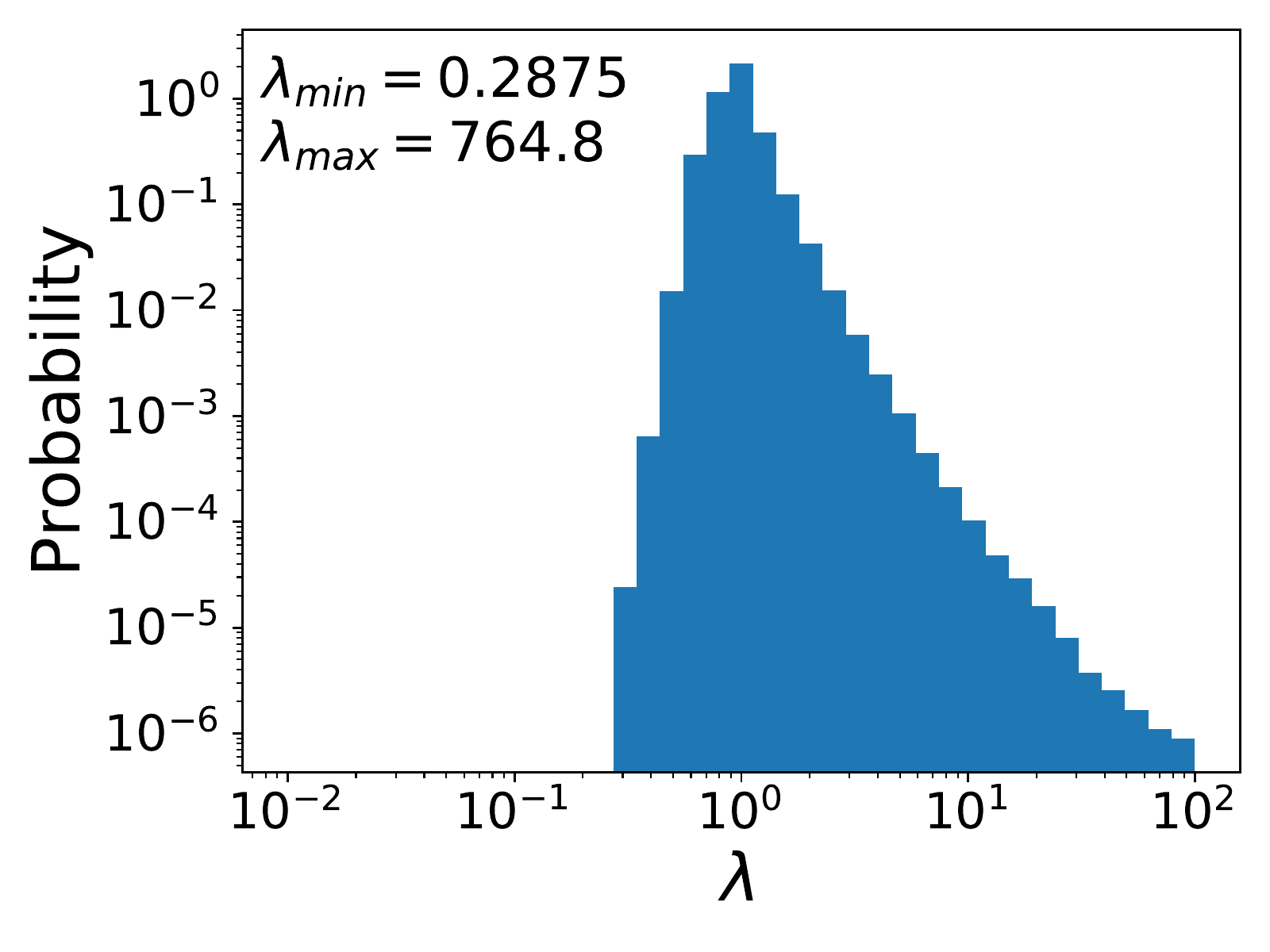}
        \label{fig:ev:both}
    }
    \caption{Histogram of the eigenvalue spectra of $M$ for the case of only attractive, only repulsive, and combined interaction sets in \cref{fig:ev:att,fig:ev:rep,fig:ev:both}, respectively. It can be observed that the spectra in \cref{fig:ev:att,fig:ev:rep} are limited, whereas the spectrum for both interactions does not exhibit such a limitation. }
    \label{fig:ev}
\end{figure}

The results on the eigenvalue spectra can be verified numerically: Additional simulations at $\alpha=\frac{1}{2}$ were performed for all three interaction cases, and the update matrix was stored periodically. We then determined the eigenvalues and plotted their frequency as histograms in \cref{fig:ev}. Indeed, we observe the bounded spectra for limited interaction sets while the system with both interactions exhibits a more balanced eigenvalue spectrum. While the spectrum for both interactions in \cref{fig:ev:both} is not entirely balanced around $1$, and hence the determinant is larger than $1$, the matrix is closer to describe an Anosov diffeomorphism and can describe one if necessary. In particular, the oscillatory strategies discussed in \cref{sec:results:both} alternate attractive and repulsive interactions to achieve eigenvalues larger and smaller than $1$ over multiple time-steps. In conclusion, we indeed observe the predicted eigenvalue spectra throughout all simulations, clearly demonstrating why limited interaction sets cannot provide mixing, whereas the combined interaction set can.

\section{Discussion} \label{sec:discussion}
In this paper, we analyzed a challenging control problem arising for active matter components, where control cannot be exerted over individual agents or particles but merely over their pairwise interaction. In particular, we attempted to find efficient microscopic mixing strategies using policy-based RL, where we observed that RL only finds efficient mixing strategies for models with combined attractive and repulsive interactions but not for those with only one type of interaction. This peculiar success/failure pattern was then analyzed using dynamical systems theory, particularly theory to hyperbolic dynamics, on the update matrix $M$ to prove that both types of interactions are required to induce mixing in arbitrary drag-dominated translational-invariant particle models. Finally, the mathematical results on $M$ were confirmed in additional training runs, demonstrating why RL could only solve the problem with both interaction types.

The results of this paper suggest further work in two primary directions: For microscopic mixing applications, strategies found in this paper could be further refined using more intermediate values of $\alpha$, a higher grid resolution, and more realistic boundary conditions and particle-particle interactions. Successful strategies in such a search could then be transferred to the experimental systems~\cite{Ross2019,Qu2020a,Qu2021b}, by implementing additional controlled interactions, possibly resulting in simpler and cheaper microfluidic mixing devices.

As a second emergent research direction, mixing could be studied in different systems: The current model could be exchanged for models that are not drag-dominated or particle-based. In particular, inertia-dominated particle systems can arise in the context of celestial mechanics and are known to display a rich mixing behavior~\cite{Stone2019}. Besides particle-based systems, continuum models can be considered both at high or low Reynolds numbers. While both of these directions are highly interesting, their computational demand will undoubtedly be larger, so that we expect a more advanced machine learning approach to be necessary.
Nevertheless, we are confident that our work paves the way for future application of RL to this area of study, namely control of complex interaction-dominated many-body systems.

The code for the simulation can be found in Ref.~\cite{Schildknecht2020}, and the OpenAI gym environment will be uploaded to GitHub and can currently be found in the supplemental material. The data is accessible at Ref.~\cite{Schildknecht2021DatasetAnonymous}.
%

\begin{ack}
    We thank Arjuna Subramanian, Jerry Wang, Pranav Bhamidipati, Ivan Jimenez, Jeremy Bernstein, Dr.\ Guannan Qu, and Dr.\ Shahriar Shadkhoo for scientific discussions and feedback on the manuscript.
    We thank Inna-Marie Strazhnik for help with the figures.
    We acknowledge funding through the Foundational Questions Institute and Fetzer Franklin Fund through FQXi 1816, the Packard Foundation (2019-69662), and the Heritage medical research institute. ANP acknowledges additional funding through the SFP SURF program.
    The authors declare no competing financial interests.
\end{ack}

\bibliography{library}

\section*{Checklist}


\begin{enumerate}

\item For all authors...
\begin{enumerate}
  \item Do the main claims made in the abstract and introduction accurately reflect the paper's contributions and scope?
    \answerYes{}
  \item Did you describe the limitations of your work?
    \answerYes{}
  \item Did you discuss any potential negative societal impacts of your work?
    \answerNA{}
  \item Have you read the ethics review guidelines and ensured that your paper conforms to them?
    \answerYes{} 
\end{enumerate}

\item If you are including theoretical results...
\begin{enumerate}
  \item Did you state the full set of assumptions of all theoretical results?
      \answerYes{See \cref{sec:analytics}.}
	\item Did you include complete proofs of all theoretical results?
        \answerYes{See \cref{sec:analytics}.}
\end{enumerate}

\item If you ran experiments...
\begin{enumerate}
  \item Did you include the code, data, and instructions needed to reproduce the main experimental results (either in the supplemental material or as a URL)?
  \answerYes{Simulation code can be found in Ref.~\cite{Schildknecht2020}, the OpenAI gym will be released on GitHub and can currently be found in the supplemental material.}
  \item Did you specify all the training details (e.g., data splits, hyperparameters, how they were chosen)?
  \answerYes{Methods were outlined in \cref{sec:methods}, a complete list of the necessary parameters can be found in the supplemental material A.}
\item Did you report error bars (e.g., with respect to the random seed after running experiments multiple times)?
\answerNo{The experiments reported, in particular the strategy screening, took considerable compute power, and cannot easily repeated with limited budget. In addition, the strategies were accessed qualitatively, so there is a secondary uncertainty involved, limiting the ability to include error bars.}
	\item Did you include the total amount of compute and the type of resources used (e.g., type of GPUs, internal cluster, or cloud provider)?
        \answerYes{See supplemental material A.3}
\end{enumerate}

\item If you are using existing assets (e.g., code, data, models) or curating/releasing new assets...
\begin{enumerate}
  \item If your work uses existing assets, did you cite the creators?
    \answerNA{}
  \item Did you mention the license of the assets?
    \answerNA{}
  \item Did you include any new assets either in the supplemental material or as a URL?
      \answerYes{All data gathered during our simulations were uploaded to an Open Access Zenodo repository~\cite{Schildknecht2021DatasetAnonymous}.}
  \item Did you discuss whether and how consent was obtained from people whose data you're using/curating?
    \answerNA{}
  \item Did you discuss whether the data you are using/curating contains personally identifiable information or offensive content?
    \answerNA{}
\end{enumerate}

\item If you used crowdsourcing or conducted research with human subjects...
\begin{enumerate}
  \item Did you include the full text of instructions given to participants and screenshots, if applicable?
    \answerNA{}
  \item Did you describe any potential participant risks, with links to Institutional Review Board (IRB) approvals, if applicable?
    \answerNA{}
  \item Did you include the estimated hourly wage paid to participants and the total amount spent on participant compensation?
    \answerNA{}
\end{enumerate}

\end{enumerate}
\appendix
\newpage
\section{Tabulated list of the parameters} \label{app:params}
While all relevant parameters are listed in the main text, we want to list all parameters used in the simulation. In particular, both the parameter set for the environment and the RL hyperparameters are listed in \cref{app:params:env,app:params:ray}, respectively.

\subsection{Environment}\label{app:params:env}
Here, we list the parameters for the simulation framework. It should be noted that several features are deactivated. In particular, there is no Brownian motion and no fluid integration taking place for this paper's purpose.
\begin{center}
\begin{tabular}{l|l|p{.3\linewidth}}
   Parameter Name                           & Value                    & Notes\\\hline 
   \verb|sweep_experiment                 | & False                    & \\
   \verb|mixing_experiment                | & True                     & \\
   \verb|run_id                           | & 0                        & Only relevant if \verb|sweep_experiment| is true.\\
   \verb|savefreq_fig                     | & 1000000                  & Never stores any figures.\\
   \verb|savefreq_data_dump               | & 100000                   & Never store data dumps.\\
   \verb|use_interpolated_fluid_velocities| & True                     & Irrelevant because fluids are deactivated by \verb|Rdrag|$=0$\\
   \verb|dt                               | & 0.05                     & \\
   \verb|T                                | & 5                        & \\
   \verb|particle_density                 | & 6.0                      & Leading to $96$ particles in a $4\times 4$ area.\\
   \verb|MAKE_VIDEO                       | & False                    & \\
   \verb|SAVEFIG                          | & False                    & \\
   \verb|const_particle_density           | & False                    & \\
   \verb|measure_one_timestep_correlator  | & False                    & \\
   \verb|periodic_boundary                | & True                     & \\
   \verb|activation_fn_type               | & \verb|activation_matrix |& \\
   \verb|L                                | & 2                        & Simulation area is in both directions from $-L$ to $L$.\\
   \verb|m_init                           | & 1.0                      & \\
   \verb|activation_decay_rate            | & 10.0                     & \\
   \verb|spring_cutoff                    | & 1.5                      & \\
   \verb|spring_k                         | & 3.0                      & \\
   \verb|spring_k_rep                     | & 3.0                      & Spring constant $k$ for the repulsive interaction is chosen as strong as for the attractive interaction.\\
   \verb|spring_r0                        | & 0                        & \\
   \verb|LJ_eps                           | & 0                        & \\
   \verb|brownian_motion_delta            | & 0                        & \\
   \verb|mu                               | & 10.0                     & \\
   \verb|Rdrag                            | & 0                        & \\
   \verb|drag_factor                      | & 1                        & \\
   \verb|spring_lower_cutoff              | & 0.015                    & \\
   \verb|n_part                           | & 96
\end{tabular}
\end{center}

\subsection{Reinforcement learning framework: \code{rl-lib}} \label{app:params:ray}
Here, we list the hyperparameters chosen for the RL framework. In particular, we used the Proximal Policy Optimization as implemented in \code{rl-lib} in conjunction with Population Based Training implemented in \code{ray}~\cite{Liaw2018}. It should be noted that while some of the parameters are initially constrained to an interval, mutations caused by PBT can go beyond the initial interval. Here, $\mathcal U([a,b])$ denotes the uniform distribution over the interval $[a,b]$, and $\mathcal I(a,b)$ is a random integer between $a$ and $b$ (inclusive).
\begin{center}
\begin{tabular}{l|l}
   Parameter Name                                         & Value                                                                                                                 \\\hline
   \verb|time_attr|                                       & \verb|time_total_s|                                                                                                   \\
   \verb|metric|                                          & \verb|episode_reward_mean|                                                                                            \\
   \verb|mode|                                            & \verb|max|                                                                                                            \\
   \verb|perturbation_interval|                           & 120                                                                                                                   \\
   \verb|hyperparam_mutations|: \verb|lambda|             & $\mathcal U([0.8,1])$                                                                                                 \\
   \verb|hyperparam_mutations|: \verb|clip_param|         & $\mathcal U([0.01,0.7])$                                                                                              \\
   \verb|hyperparam_mutations|: \verb|lr|                 & $[ 1\cdot 10^{-2}, 5\cdot 10^{-3}, 1\cdot 10^{-3}, 5\cdot 10^{-4},$\\&$~ 1\cdot 10^{-4}, 5\cdot 10^{-5}, 1\cdot 10^{-5} ]$  \\
   \verb|hyperparam_mutations|: \verb|num_sgd_iter|       & $\mathcal I(1,30)$                                                                                                    \\
   \verb|hyperparam_mutations|: \verb|sgd_minibatch_size| & $\mathcal I(128,16384)$                                                                                               \\
   \verb|hyperparam_mutations|: \verb|train_batch_size|   & $\mathcal I(2000,60000)$                                                                                              
\end{tabular}
\end{center}

\subsection{Computing resources used} \label{app:params:resources}
Here, we list the amount of computing resources used for our simulations. There were two sets of simulations, one to determine the strategies, the second one extracting the update matrix $M$. For both of these, $\{a,b,c, \cdots\}$ means that the experiment was repeated for all values in this set. If multiple sets are listed, simulations were performed for all different combinations. All other numbers are for a single combination. For example, every combination of a specific $\alpha$ and interaction set used 16 CPU cores. The simulation parameters for the strategy screen (see Sec.~4) are:
\begin{center}
\begin{tabular}{p{.45\linewidth}|p{.45\linewidth}}
   Parameter Name                                                    & Value                                                                                                        \\ \hline
   $\alpha$                                                          & $\{0, \frac{1}{8},\frac{2}{8},\frac{3}{8},\frac{4}{8},\frac{5}{8},\frac{6}{8},\frac{6}{8},\frac{7}{8}, 1 \}$ \\
   (attractive interactions allowed, repulsive interactions allowed) & \{(\verb|True|, \verb|True|), (\verb|True|, \verb|False|), (\verb|False|, \verb|True|) \}                \\
   Number of CPU cores                                               & 16                                                                                                           \\
   Number of GPUs                                                    & 1                                                                                                            \\
   Training time                                                     & 2 or 3 days, depending if the 2 days run converged completely                                                \\
   Type of Resource                                                  & University cluster
\end{tabular}
\end{center}

The simulation parameters for the eigenvalue screen are:
\begin{center}
\begin{tabular}{p{.45\linewidth}|p{.45\linewidth}}
   Parameter Name                                                    & Value                                                                                                        \\ \hline
   $\alpha$                                                          & $\frac{1}{2}$ \\
   (attractive interactions allowed, repulsive interactions allowed) & \{(\verb|True|, \verb|True|), (\verb|True|, \verb|False|), (\verb|False|, \verb|True|) \}                \\
   Number of CPU cores                                               & 16                                                                                                           \\
   Number of GPUs                                                    & 1                                                                                                            \\
   Training time                                                     & 3 days\\
   Type of Resource                                                  & University cluster
\end{tabular}
\end{center}

\newpage
\section{Curated time series for each strategy} \label{app:curated-strategies}
\begin{figure}[h]
    \centering
    \subfloat[][collapse all]{
        \timeseries{ca}{005}{010}{015}{020}
        \label{fig:curated:att:ca}
    }\\
    \subfloat[][collapse all, careful]{
        \timeseries{cac}{000}{033}{067}{100}
        \label{fig:curated:att:cac}
    }\\
    \subfloat[][collapse some]{
        \timeseries{cs}{000}{015}{030}{045}
        \label{fig:curated:att:cs}
    }\\
    \subfloat[][activate little]{
        \timeseries{al}{000}{033}{067}{100}
        \label{fig:curated:att:al}
    }\\
    \caption{Curated examples for all strategies emerging in attractive-interactions-only simulations. It should be noted that \cref{fig:curated:att:ca} is the same time series as depicted in the main manuscript in Fig.~2(a).}
    \label{fig:curated:att}
\end{figure}

\begin{figure}
    \centering
    \subfloat[][repulsive spreading]{
        \timeseries{rs}{000}{015}{030}{045}
        \label{fig:curated:rep:rs}
    }\\
    \subfloat[][activate one side]{
        \timeseries{aos}{000}{010}{020}{030}
        \label{fig:curated:rep:aos}
    }\\
    \caption{Curated examples for all strategies emerging in repulsive-interactions-only simulations. It should be noted that \cref{fig:curated:rep:aos} is the same time series as depicted in the main manuscript in Fig.~3(a).}
    \label{fig:curated:rep}
\end{figure}

\begin{figure}
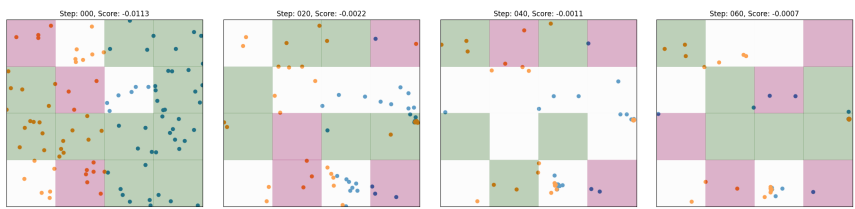
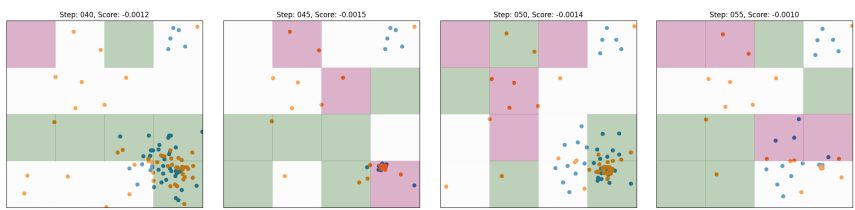
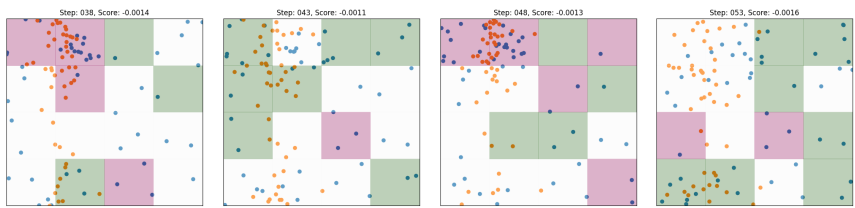
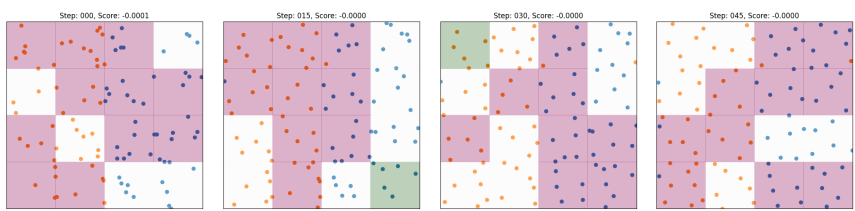

    \centering
    \subfloat[][collapse all (with both interactions allowed)]{
        \timeseries{ca-both}{000}{020}{040}{060}
        \label{fig:curated:both:ca}
    }\\
    \subfloat[][oscillation w/ collapse]{
        \timeseries{owc}{040}{045}{050}{055}
        \label{fig:curated:both:owc}
    }\\
    \subfloat[][oscillation w/o collapse]{
        \timeseries{osc}{038}{043}{048}{053}
        \label{fig:curated:both:osc}
    }\\
    \subfloat[][attractive-repulsive spreading]{
        \timeseries{ars}{000}{015}{030}{045}
        \label{fig:curated:both:ars}
    }\\
    \caption{Curated examples for all strategies emerging if attractive and repulsive interactions are available. It should be noted that \cref{fig:curated:both:owc,fig:curated:both:osc} are the same time series as depicted in the main manuscript in Fig.~4(a) and (b), respectively.}
    \label{fig:curated:both}
\end{figure}
\end{document}